\documentclass{article}

\usepackage{hyperref}

\usepackage{url}
\usepackage{amsmath}
\usepackage{bbold}
\usepackage{amssymb}
\usepackage{acronym} 
\usepackage{booktabs} 
\usepackage[separate-uncertainty]{siunitx} 
\usepackage{array} 
\usepackage{array,multirow,graphicx}
\usepackage{algorithm}
\usepackage{algorithmic}
\usepackage{soul}
\usepackage{xcolor,colortbl}
\usepackage{tikz}
\usepackage{amsthm}
\newcommand{\dittoarrow}{{\hspace*{0.7ex}\raisebox{0.5ex}{\tikz{\draw[->] (0,0.1) |- (0.3,0);}}}}

\usepackage{etoolbox}
\robustify\itshape
\robustify\bfseries

\DeclareMathAlphabet{\mathbbb}{U}{bbold}{m}{n}
\DeclareSymbolFontAlphabet{\mathbb}{AMSb}

\usepackage{enumitem}

\newlist{questions}{enumerate}{2}
\setlist[questions,1]{label=\textbf{Q\arabic*.},ref=Q\arabic*}
\setlist[questions,2]{label=(\alph*),ref=\thequestionsi(\alph*)}

 \usepackage[accepted]{icml2021}

\icmltitlerunning{Exploring Adversarial Robustness of Deep Metric Learning}

\acrodef{ML}[ML]{machine learning}
\acrodef{DNN}[DNN]{deep neural network}
\acrodefplural{DNN}[DNNs]{deep neural networks}
\acrodef{DML}[DML]{Deep Metric Learning}
\acrodef{FGSM}[FGSM]{Fast Gradient Sign Method}
\acrodef{PGD}[PGD]{Projected Gradient Decent}
\acrodef{CW}[C\&W]{Carlini-Wagner}

\DeclareMathOperator*{\argmin}{arg\,min}

\newcommand{\urlrepository}{(anonymized repository) \url{https://github.com/anonymous-koala-supporter/adversarial-deep-metric-learning}}
\newcommand{\urlgallery}{(anonymized gallery) \url{https://anonymous-koala-supporter.github.io/sample-gallery/}}

\DeclareMathOperator{\bfx}{{\bf x}}
\DeclareMathOperator{\bfy}{{\bf y}}
\DeclareMathOperator{\bfz}{{\bf z}}
\DeclareMathOperator{\bfa}{{\bf a}}
\DeclareMathOperator{\calD}{{\mathcal D}}
\DeclareMathOperator{\calX}{{\mathcal X}}
\DeclareMathOperator{\calY}{{\mathcal Y}}

\newtheorem{theorem}{Theorem}
\newtheorem{lemma}{Lemma}

\newcommand{\datadist}{\mathcal{D}}
\newcommand{\xset}{\mathcal{X}}
\newcommand{\yset}{\mathcal{Y}}
\newcommand{\hyposet}{\mathcal{H}}

\newcommand{\datapoint}{\mathbf{x}}
\newcommand{\dataset}{S}

\newcommand{\param}{\theta}
\newcommand{\paramset}{\Theta}

\newcommand{\sphere}{S^{d}}
\newcommand{\distfunc}{d_{\param}}
\newcommand{\lossfunc}{l}
\newcommand{\norm}{\ell}

\newcommand{\class}{c}
\newcommand{\samplepoint}{\mathbf{z}}

\newcommand{\knnfunc}{k}

\newcommand{\anchorset}{A}
\newcommand{\anchor}{\mathbf{a}}

\newcommand{\highlight}{\bfseries}
\def\Vhrulefill{\leavevmode\leaders\hrule height 0.7ex depth \dimexpr0.4pt-0.7ex\hfill\kern0pt}
\newcommand{\notapp}{\Vhrulefill{}\text{N/A}\Vhrulefill{}}
\newcommand{\hlrow}{\rowcolor{gray!15}}

\newcommand{\myparagraph}[1]{\vspace*{-1.5ex}\paragraph{#1}}

\begin{document}

\twocolumn[
\icmltitle{Exploring Adversarial Robustness of Deep Metric Learning}



\icmlsetsymbol{equal}{*}

\begin{icmlauthorlist}
\icmlauthor{Thomas Kobber Panum}{aau}
\icmlauthor{Zi Wang}{uwm}
\icmlauthor{Pengyu Kan}{uwm}
\icmlauthor{Earlence Fernandes}{uwm}
\icmlauthor{Somesh Jha}{uwm}
\end{icmlauthorlist}

\icmlaffiliation{uwm}{Department of Computer Sciences, University of Wisconsin-Madison, Wisconsin, USA}
\icmlaffiliation{aau}{Department of Electronic Systems, Aalborg University, Denmark}

\icmlcorrespondingauthor{Thomas Kobber Panum}{tkp@es.aau.dk}

\icmlkeywords{deep metric learning, adversarial robustness}

\vskip 0.3in
]


\printAffiliationsAndNotice{}  

\begin{abstract}

  Deep Metric Learning (DML), a widely-used technique, involves learning a distance metric between pairs of samples. DML uses deep neural architectures to learn semantic embeddings of the input, where the distance between similar examples is small while dissimilar ones are far apart. Although the underlying neural networks produce good accuracy on naturally occurring samples, they are vulnerable to adversarially-perturbed samples that reduce performance. We take a first step towards training robust DML models and tackle the primary challenge of the metric losses being dependent on the samples in a mini-batch, unlike standard losses that only depend on the specific input-output pair. We analyze this dependence effect and contribute a robust optimization formulation. Using experiments on three commonly-used DML datasets, we demonstrate \numrange[range-phrase = --]{5}{76} fold increases in adversarial accuracy, and outperform an existing DML model that sought out to be robust.

\end{abstract}
\section{Introduction}\label{sec:intro}
Many machine learning (ML) tasks rely on ranking entities based on the similarities of data points in the same class.
\ac{DML} is a popular technique for such tasks, particularly for applications involving test-time inference of classes that are not present during training (e.g. zero-shot learning).
Example applications of \ac{DML} include person re-identification~\citep{hermans17}, face verification~\citep{schroff15,deng19}, phishing detection~\citep{abdelnabi20}, and image retrieval~\citep{wu17,roth19}.
At its core, \ac{DML} relies on state-of-the-art deep learning techniques for training models that output lower-dimensional semantic feature embeddings from high-dimensional inputs.
Points in this embedding space cluster similar inputs together while dissimilar inputs are far apart.

\begin{figure*}[h]
  \includegraphics[width=\textwidth]{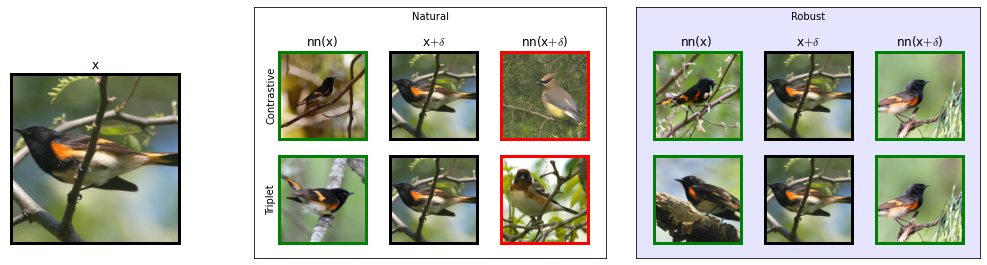}
  \caption{
    \label{fig:exampleinfer}
    Example of inference of a naturally-trained \ac{DML} model and robustly trained variant on the CUB200-2011 dataset.
    Each model infers the class of the natural data point $\datapoint$, and its perturbed $\datapoint + \delta$ counterpart, using the class of the nearest anchor $nn(\cdot)$.
    Green and red borders indicate correct- (same class) and incorrect inference, respectively.
    Both models infer the natural input correctly, however, the naturally-trained \ac{DML} model fails to infer the adversarial perturbed input correctly.
  }
\end{figure*}

Traditional deep learning classifiers are vulnerable to adversarial examples~\citep{szegedy14,biggio13} --- inconspicuous input changes that can cause the model to output attacker-desired values.
Few studies have addressed whether \ac{DML} models are similarly susceptible towards these attacks, and the results are contradictory~\cite{abdelnabi20,panum20}.
Given the wide usage of \ac{DML} models in diverse ML tasks, including security-oriented ones, it is important to clarify their susceptibility towards attacks and ultimately address their lack of robustness.
We investigate the vulnerability of \ac{DML} towards these attacks and address the open problem of training DML models using robust optimization techniques~\citep{bental09,madry18}.

A key challenge in robust training of DML models concerns the so-called \emph{metric losses}~\citep{wu17,wang19,chechik10,schroff15}.
Unlike loss functions used in typical deep learning settings, the metric loss for a single data point is interdependent on other data points.
For example, the widely-used triplet loss requires three input points: an anchor, a positive sample similar to the anchor, and a negative sample dissimilar to the anchor.
For training, this interdependence impacts the effectiveness of learning~\cite{wu17}. Thus, several works have identified sampling strategies that turn mini-batches into tuples or triplets to ensure that training remains effective~\cite{schroff15,yuan17,xuan20}


This \emph{interdependence} between data points of metric losses poses a challenge for creating effective adversarial perturbations, as these are typically computed by approximating the inconspicuous noise that maximizes the loss for the specific data point.
Consequently, as adversarial training depends on this ability during training, it has to remain efficient in order to reduce the additional computation as natural training procedures for certain \ac{DML} models are already considered resource intensive~\cite{roth20}.
Additionally, metric losses are sensitive to samples with high levels of noise during training, that can cause training to reach an undesired local minima~\cite{wu17}.
Adversarial perturbations are effectively noise, and thus adversarial training procedure for \ac{DML} models has to account for this sensitivity.


We systematically approach the above challenges and contribute a robust training objective formulation for DML models by considering the two widely-used metric losses --- contrastive and triplet loss.
An example of the influence the robust training objective on inference is shown in Figure~\ref{fig:exampleinfer}.
Our key insight is that during an inference-time attack, adversaries seek to perturb data points such that the intra-class distance maximize, and thus this behavior needs to be accounted for during training to improve robustness.
Recent work has attempted to train robust \ac{DML} models but has not considered the dependence and sensitivity to sampling~\citep{abdelnabi20}.
When we subject these models to our attack techniques, we find that their robustness is actually less than what is reported.

Prior work on traditional classifiers have established a connection between Lipschitz constant and robustness~\cite{hein17}.
Our intuition is the adversarial training of lead to a lower Lipschitz constant of the deep metric embedding.
We explore this further in supplementary materials.

\textbf{Contributions.}
\begin{itemize}
\item We contribute a principled robust training framework for \ac{DML} models by considering the dependence of metric losses on the other data points in the mini-batch and the sensitivity to sampling.
\item We experiment with naturally-trained DML models across three commonly-used datasets for \ac{DML} (CUB200-2011, CARS196, SOP) and show that they have poor robustness --- their accuracy (R@1) drops from \num{59.1}\% (or more) to \num{4.0}\% (or less) when subjected to PGD attacks of the proposed attack formulation.
\item Using our formulation for adversarial training, \ac{DML} models reliably increase their adversarial robustness, outperforming prior work.
  For $\norm_{\infty}(\epsilon = 0.01)$, we obtain an adversarial accuracy of \num{53.6}\% compared to the state-of-the-art natural accuracy baseline of \num{71.8}\% for the SOP dataset (in terms of R@1 score, a common metric in DML to assess the accuracy of models).
  Furthermore, the resulting robust model accuracies are largely unaffected for natural (unperturbed) samples.
\end{itemize}
\section{Related Work}

\paragraph{Deep Metric Learning.} \ac{DML} is a popular technique to obtain semantic feature embeddings with the property that similar inputs are geometrically close to each other in the embedding space while dissimilar inputs are far apart~\cite{roth20}. \ac{DML} losses involve pairwise distances between embeddings~\cite{boudiaf20}.
Examples include contrastive loss~\cite{hadsell06}, triplet loss~\cite{schroff15}, Neighborhood Component Analysis~\cite{goldberger04}, and various extensions of these losses~\cite{sohn16,wang19,zheng19}.
Throughout this work, we refer to these types of losses as \emph{metric losses}.
Recent surveys~\cite{roth20,musgrave20} highlight that performance of newer metric losses are lesser than previously reported.
Thus, we focus on the two established metric losses --- contrastive and triplet --- as they are widely used and have good performance.

\paragraph{Adversarial Robustness.} Since early work in the ML community discovered adversarial examples in deep learning models~\cite{szegedy14,biggio13}, a big focus has been to train adversarially-robust models.
We focus on robust optimization-based training that utilizes a saddle-point formulation (min-max)~\cite{bental09,madry18}.
To the best of our knowledge, training \ac{DML} models using robust-optimization techniques has not been thoroughly studied, and only recently has work begun in this area~\cite{abdelnabi20}.

Using the Generative Adversarial Network architecture~\cite{goodfellow14gan}, \citet{duan18} create a framework that uses generative models during training to derive hard negative samples from easy negatives. They focus on improving the effectiveness of \emph{naturally} training \ac{DML} models rather than obtaining adversarial robustness, which is our focus.

Recent studies have shown that metric losses can function as a supplementary regularization method that enhances adversarial robustness of deep neural network classifiers (e.g., CNNs)~\cite{mao19,li19}. However, these studies are not applicable to training robust \ac{DML} models, as they do not solve the problem of dependence between data points due to the use of metric losses. We propose a principled framework for robustly training \ac{DML} models that accounts for this problem.

\section{Towards Robust Deep Metric Models}
\label{sec:background}
First, we describe some basic machine learning (ML) notation and concepts required to describe our algorithm.  We assume a data distribution $\datadist$ over $\xset \times \yset$, where $\xset$ is the sample space and $\yset = \{ y_1,\cdots,y_L \}$ is the finite space of labels.
Let
$\datadist_{\xset}$ be the marginal distribution over $\xset$ induced by $\datadist$~\footnote{The measure of set $Z \subseteq \xset$ in distribution $\datadist_{\xset}$ is the measure of the set $Z \times \yset$ in distribution $\datadist$.}.
Given $Y \subseteq \yset$ we define $\datadist_Y$ to be the measure of the subsets of $\xset \times Y $ induced by $\datadist$.
For $y \in \datadist$, $\datadist_y$ and $\datadist_{-y}$ denote the measures $D_{ \{ y \} }$ and $D_{\yset \setminus \{ y \} }$, respectively.

In the {\it empirical risk minimization (ERM)} framework we wish to solve the following optimization problem:
\begin{equation}
  \min_{w \in \hyposet} \; E_{(\datapoint,y) \sim \datadist} \; \lossfunc(w,\datapoint,y)
\end{equation}
In the equation given above $\hyposet$ is the hypothesis space and
$l$ is the loss function. We will denote vectors in boldface (e.g. $\datapoint$, $\bfy$). Since the distribution is usually unknown, a learner solves the following problem over a data set $S=\{ (\datapoint_1,y_1), \cdots. (\datapoint_n,y_n) \}$ sampled from the distribution $\datadist$.
\begin{equation}
  \min_{ w \in \hyposet } \; \frac{1}{n} \sum_{i=1}^n  \lossfunc(w,\datapoint_i,y_i)
\end{equation}
Once we have solved the optimization problem given above, we obtain a $w^* \in \hyposet $ which yields a classifier $F \colon \xset \rightarrow \yset$ (the classifier is usually parameterized by $w^*$, but we will omit it for brevity).

\subsection{Deep Metric Models}

The goal of {\it deep metric learning (DML)} is to create a
{\it deep metric model} $f_\param$ is function from $\xset$ to $\sphere$, where $\param \in \paramset$ is a parameter and $\sphere$ is an unit sphere in $\mathbb{R}^d$ (i.e.  $\datapoint \in \sphere$ iff $\| \datapoint \|_2 \; = \; 1$).
Since deep metric models embed a space $\xset$ (which can itself be a metric space) in another metric space, we also sometimes refer to them deep embedding.
Frequently, deep metric models use very different loss functions than typical classification networks described previously.
Next we discuss two kinds of loss functions -- contrastive and triplet.
Let $S= \{(\datapoint_1,y_1),\cdots,(\datapoint_n,y_n) \}$ be a dataset drawn from $\datadist$.
A {\it contrastive} loss function $l_c (\param,(\datapoint,y)$, $(\datapoint_1,y_1))$ of labeled samples from $\xset \times \yset$ and is defined as:
\begin{equation}
  1_{y = y_1} \; \distfunc (\datapoint,\datapoint_1) + 1_{y \not= y_1} \; [\alpha - \distfunc (\datapoint,\datapoint_1)]
\end{equation}
In the equation given above, $1_E$ is an indicator function for event $E$ ($1$ if event $E$ is true and $0$ otherwise), and $\distfunc (\datapoint,\datapoint_1)$ is $\sqrt{ \sum_{j=1}^d (f_\param (\datapoint)_j -
f_\param(\datapoint_1)_j)^2}$, the $\norm_2$ distance in the embedding space.
The goal of the contrastive loss function is to reduce the distance
in the embedding space between
two samples with the same label, and analogously
increase the distance in the embedding space between the two samples with different labels.
A  {\it triplet} loss function $\lossfunc_t$ is defined over three $\lossfunc_t (\param,(\datapoint,y),(\datapoint_1,y_1),(\datapoint_2,y_2))$ labeled samples and is defined as follows:

\begin{equation}
  1_{y = y_1} \; 1_{y \not= y_2} \; [ \distfunc (\datapoint,\datapoint_1) - \distfunc (\datapoint,\datapoint_2) + \alpha ]_+
\end{equation}

In the equation given above $[ x ]_+$ is $\max (x,0)$. In order for the
expression to be non-zero $(\datapoint_1,y_1)$ has to have the
same label as $(\datapoint,y)$, and $(\datapoint_2,y_2)$ has to have a different
label as $(\datapoint,y)$.

\subsection{Attacks on Deep Metric Models}\label{sec:attackdml}
Assume that we have learned a deep embedding network with parameter $\param \in \Theta$ using one of the loss functions described above.
Next we describe how the network is used. Let $A = \{ (\anchor_1,\class_1),\cdots, (\anchor_m,\class_m)\}$ be a reference or test dataset (e.g. a set of faces along with their label).
$A$ is distinct from the dataset $S$ used during training time.
Suppose we have a sample $\samplepoint$ and let $k (\anchorset,\samplepoint)$ be the index that corresponds to $\argmin_{j \in \{1,\cdots,m \} } \distfunc (\anchor_j,\samplepoint)$\footnote{In case one or more anchors share the minimal distance to $\samplepoint$, the tie is broke by a random selection among these anchors.}.
We predict the label of $\samplepoint$ as $lb(\anchorset,\samplepoint) \; = \; \class_{\knnfunc(\anchorset,\samplepoint)}$(we will use the functions $\knnfunc(.,.)$ and $lb(.,.)$ throughout this section).

Next we describe test-time attacks on a deep embedding with parameter $\param$.
Let $\samplepoint \in \xset$.
{\it Untargeted attack} on $\samplepoint$ can be described as follows (we want the perturbed point to have a different label than before):
\begin{equation}\label{eq:untargeted}
\begin{array}{l}
  \min_{\delta \in \xset} \; \mu (\delta) \;\\
  \mbox{\it such that} \; lb(\anchorset,\samplepoint) \not= lb(\anchorset,\samplepoint+\delta)
\end{array}
\end{equation}

{\it Targeted attack} (with a target label $t \not= lb(A,\samplepoint)$) can be described as follows (we desire to the predicted label of the perturbed point to be a specific label):
\begin{equation}
\begin{array}{l}
  \min_{\delta \in \xset} \; \mu (\delta) \;\\
  \mbox{\it such that} \; lb (A,\samplepoint+\delta) = t
\end{array}
\end{equation}
In the formulations given above we assume that $\xset$ is a metric space with $\mu$ a metric on $\xset$ (e.g. $\xset$ could $\mathbb{R}^{n}$ with usual norms, such as $\norm_\infty$, $\norm_1$, or $\norm_p$ (for $p \geq 2$)).
Any algorithm that solves the optimization problem described above leads to a specific attack on deep metric models.

\subsection{Robust Deep Metric Models}~\label{sec:robustdml}

Let $S = \{ (\datapoint_1,y_1),\cdots, (\datapoint_n,y_n) \}$ be a dataset drawn from distribution $\datadist$.
For a sample $(\datapoint_i,y_i)$ where $1 \leq i \leq n$ we define the following surrogate loss function $\hat{\lossfunc} (\param,(x_i,y_i),S)$ for the contrastive loss function $l_c$ :
\begin{equation}
  \hat{\lossfunc} (\param,(x_i,y_i),S) = \frac{1}{n} \sum_{j=1}^n \lossfunc_c (\param,(\datapoint_i,y_i),(\datapoint_j,y_j))
\end{equation}
Similarly, for the triplet loss function $l_t$ we can define the following surrogate loss function $\hat{\lossfunc} (\param,(x_i,y_i),S)$:
\begin{equation}
  \frac{1}{n_{y_i} n^{-}_{y_i}} \sum_{j=1}^{n_{y_i}}
  \sum_{k=1}^{n^{-}_{y_i}}  \lossfunc_t (\param,(\datapoint_i,y_i),(\datapoint_j,y_j), (\datapoint_k,y_k))
\end{equation}
Let $S_y$ and $S_{-y}$ be defined as the following sets: $\{ (\datapoint,y) \; \mid \; (\datapoint,y) \in S \}$ and $\{ (\datapoint,y') \; \mid \; (\datapoint,y') \in S \; \mbox{and} \; y' \not= y \}$.
In the equation given above the sizes of the sets $S_y$ and $S_{-y}$ are denoted by $n_y$ and $n^-_y$, respectively.

Having defined the surrogate loss function $\hat{\lossfunc}$ the learner's problem can be defined as:

\begin{equation}
\min_{\param \in \paramset} \frac{1}{n} \sum_{i=1}^n \hat{\lossfunc}(\param,(\datapoint_i,y_i),S)
\end{equation}
Recall that the learner's problem for the usual classification case is:
\begin{equation}
\min_{ w \in \hyposet } \; \frac{1}{n} \sum_{i=1}^n  \lossfunc(w,\datapoint_i,y_i)
\end{equation}
Note that in the classification case the loss function $\lossfunc(w,\datapoint_i,y_i)$  of a sample $(\datapoint_i,y_i)$ does not depend on the other samples in the dataset $S$.
However, in the deep metric model case the surrogate loss function $\hat{\lossfunc}(\param,(\datapoint_i,y_i),S)$ for a sample $(\datapoint_i,y_i)$ depends on the rest of the data set $S$ (see the equations for $\hat{\lossfunc}$)
This is the main difference between the embedding and classification scenarios.

\noindent
{\it Formulation 1.} Let $B_p (\datapoint,\epsilon)$ denote the $\epsilon$-ball around the sample $\datapoint$ using the $\norm_p$-norm. The straightforward robust formulation is given in the equation below.
\begin{equation}
\min_{\param \in \paramset} \max_{(\samplepoint_1,\cdots,\samplepoint_n) \in \prod_{j=1}^n B_p (\datapoint_j,\epsilon) } \; \;
\frac{1}{n} \sum_{i=1}^n \hat{\lossfunc}(\param,(\samplepoint_i,y_i),S)
\end{equation}
In the formulation given above, all samples are adversarially perturbed at the same time (note that the $\max$ is outside the summation).
Therefore, this formulation is not convenient for current training algorithms, such as SGD and ADAM. This is because the entire dataset $S$ has to be perturbed at the same time. Moreover, this formulation is not conducive to various sampling strategies used in training of deep metric models.

\noindent
{\it Formulation 2.}
In this formulation we push the $\max$ inside the sum so that each term can be individually
processed. This is especially useful for adversarial training because each tuple or triple
can be perturbed separately. Our formulation will be indexed by $r$ ($r \in \{ 1,2 \}$
for contrastive loss and $r \in \{1,2,3 \}$ for triplet loss). Intuitively, $r$ denotes
what component of the tuple of triple is being perturbed. We define operator $\max (r,\epsilon,\param)$ which perturbs
the $r$-th component in an $\epsilon$ ball to maximize the loss. For example, $\max (r,\epsilon,\param)$
for $((\bfx,y),(\bfx_1,y_1))$ is defined as:
\begin{equation}
  \max_{\bfz \in B_p (\bfx_1,\epsilon)} l_c (\param,(\bfx,y),(\bfz,y_1))
\end{equation}
Now we can define $\hat{\lossfunc_r}$ for the contrastive case as:
\begin{equation*}
  \hat{\lossfunc_r} (\param,(x_i,y_i),S) = \frac{1}{n} \sum_{j=1}^n \; \max(r,\epsilon,\param) ((\datapoint_i,y_i),(\datapoint_j,y_j))
\end{equation*}
The equation for the triplet loss is similar. Now the entire minimization problem becomes.
\begin{equation}
\min_{\param \in \paramset} \frac{1}{n} \sum_{i=1}^n
\hat{\lossfunc_r}(\param,(\samplepoint,y_i),S)
\end{equation}

\noindent
{\it Formulation 3.} Our third formulation adds a regularizer which enforces the following informal constraint: if $\datapoint$ changes a bit, the distance in the embedding space does not change too much.
\begin{equation}
\min_{\param \in \paramset} \frac{1}{n} \sum_{i=1}^n [
\hat{\lossfunc}(\param,(\datapoint_i,y_i),S) \; + \; \lambda \max_{\samplepoint \in B_p (\datapoint_i,\epsilon)} \distfunc(\samplepoint,\datapoint_i) ]
\end{equation}

These robust optimization formulations follow the common notion of robustness from robust optimization~\citep{bental09}, thus given an algorithm for solving one of the robust optimization formulations, leads to a robust model.

\subsection{Attack Algorithm}~\label{sec:attackalg}

We will focus on untargeted attacks because our main goal is to use
these algorithms to robustify embeddings using adversarial training.
Recall that $d_\theta (\bfx,\bfx_1)$ is the $l_2$ distance between
$f_\theta (\bfx)$ and $f_\theta (\bfx_1)$.  The gradient
$\nabla_{\bfx} d_\theta (\bfx,\bfx_1 )$ of $d_\theta
(\bfx,\bfx_1)$ with-respect-to (wrt) to $\bfx$ is given by:
\begin{equation}
  \frac{1}{ d_\theta (\bfx,\bfx_1 )} (f_\theta (\bfx)-f_\theta(\bfx_1))^T
  \cdot \nabla_{\bfx} f_\theta (\bfx)
\end{equation}

A similar expression can be written for $\nabla_{\bfx_1} d_\theta (\bfx,\bfx_1 )$.

Consider the contrastive loss $l_c$ on a tuple $(\bfx,\bfx_1)$.
\begin{equation}
  1_{y = y_1} \; d_\theta (\bfx,\bfx_1) + 1_{y \not= y_1} \; [\alpha - d_\theta (\bfx,\bfx_1)]
\end{equation}
The gradient of the contrastive loss $\nabla_{\bfx_1} l_c (\theta,(\bfx,y),(\bfx_1,y_1))$
wrt $\bfx_1$ is shown below:
\begin{equation}
  1_{y = y_1} \; \nabla_{\bfx_1} d_\theta (\bfx,\bfx_1) - 1_{y \not= y_1} \; \nabla_{\bfx_1} d_\theta (\bfx,\bfx_1)
\end{equation}

Similar to contrastive loss, we can define gradients of
$l_t (\theta,(\bfx,y),(\bfx_1,y_1),(\bfx_2,y_2))$ wrt $\bfx$, $\bfx_1$, or $\bfx_2$.

Once we can compute the gradients of the loss, we can readily adapt attack algorithm, such as FGSM and PGD,
to our context. Note that for formulation 3 we need to only compute the
gradient of
In fact any attack algorithm that only relies on gradients of the loss function can be $d_\theta (\bfz,\bfx_1)$ with respect to $\bfx$.
adapted for our case. For example the PGD attack can be adapted for contrastive loss $l_c$ as follows:
\begin{eqnarray*}
  \bfx_1^{t+1} & = & \Pi_{\bfx_1 + B_p (\epsilon)} (\bfx_1^t + \alpha \; \nabla_{\bfx_1} l_c (\theta,(\bfx,y),(\bfx_1,y_1)) )
\end{eqnarray*}
In the equation we are showing one iteration of the PGD and $B_p (\epsilon)$ is the $\epsilon$ ball centered at the
origin using the $l_p$ norm. For computational reasons, in our attack algorithms we only perturb one of the components for the tuples of
triples.

\subsection{Adversarial Training}~\label{sec:advtrain}

Once we have the attack algorithm, adversarial training for robustifying the
model is relatively straightforward.  We assume that the attack
algorithm only perturbs one component of the tuple or triple.  Let
$\mathcal{A}^c_r (\cdot,\cdot)$ ($r \in \{ 1,2 \}$) and
$\mathcal{A}^t_r (\cdot,\cdot,\cdot)$ ($r \in \{ 1,2,3 \}$) be the
attack algorithms for the contrastive and the triple losses,
respectively. In the attack algorithms given above $r$ refers to the
index of the component being perturbed (e.g. $\mathcal{A}_2
((\bfx,y),(\bfx_1,y_1))$ and returns $((\bfx,y),(\bfx_1 + \delta
,y_1))$. Next we describe adversarial training for contrasitive loss
(the case for triple loss is similar). $\mathcal{A} (\datapoint)$ corresponds
to formulation 3 (attempts to solve $\max_{\samplepoint \in B_p (\datapoint,\epsilon)} \distfunc(\samplepoint,\datapoint))$.

As pointed before, formulation 1 is computationally prohibitive. We will
focus on formulations 2 and 3.  Let $S = \{ (\bfx_1,y_1),\cdots, (\bfx_n,y_n) \}$ be the dataset.  At each iteration, a tuple $T=((\bfx_i,y_i),(\bfx_j,y_j))$ is sampled from $S$. We construct the tuple $T'$ from $T$ using attack algorithm
(i.e. $T'= A^c_r (T)$ or $T' = A^t_r (T)$), and run one step of the learning algorithm, such as SGD or ADAM, on $T'$. This corresponds to formulation 2.
For formulation 3 we use attack algorithm $\mathcal{A}$.
\section{Experiments}\label{sec:experiments}
\begin{table*}[h!]
  \caption{
    Performance of \ac{DML} models across datasets, types of perturbations, and attack algorithms.
    Results are an average of five random seeds and best performances for a given combination of input data (\emph{Benign}, $\norm_{2}(\epsilon = 4)$, $\norm_{\infty}(\epsilon = 0.01)$) and dataset is marked in \textbf{bold}.
    We highlight our robust training technique using a shaded background and suffix them by their trained norm (e.g. ($\norm_{\infty}$)).
    Recall that, R@1 reflects a model's inference accuracy, while mAP@R reflects its ability to rank similar entities.
    Naturally-trained \ac{DML} models (Contrastive, Triplet) demonstrate low robustness towards the proposed attack formulation across the adversarial settings ($\norm_{2}(\epsilon = 4)$, $\norm_{\infty}(\epsilon = 0.01)$).
    Our robust training objective improves robustness towards adversarial attacks, and outperforms previous attempts at establishing robust \ac{DML} models (VisualPhishNet).
  }
  \label{table:comparisons}
  \vskip 0.15in
  \setlength{\tabcolsep}{4pt}
  \sisetup{detect-all = true}
  \small
  \centering
  \begin{tabular}{
    l
    l
    *{4}{
    @{\hskip 2\tabcolsep}
    S[table-format=2.2(2),table-column-width=4.4em,table-align-uncertainty = false]
    S[table-format=2.2(2),table-column-width=4.4em,table-align-uncertainty = false]
    }}
    \toprule
    & & \multicolumn{2}{c}{\textbf{CUB200-2011}} & \multicolumn{2}{@{}c@{\hskip 4\tabcolsep}}{\textbf{CARS196}} & \multicolumn{2}{@{}c@{}}{\textbf{SOP}} & \multicolumn{2}{@{}c@{}}{\textbf{VisualPhish}} \\
    \cmidrule(lr{\dimexpr 4\tabcolsep-0.5em}){3-4} \cmidrule(l{-0.5em}r{\dimexpr 4\tabcolsep-0.5em}){5-6} \cmidrule(l{-0.5em}r{\dimexpr 4\tabcolsep-0.5em}){7-8} \cmidrule(l{-0.5em}r{0.5em}){9-10}
    Model & Attack & {R@1} & {mAP@R} & {R@1} & {mAP@R} & {R@1} & {mAP@R} & {R@1} & {mAP@R}\\
    \midrule
    \multicolumn{10}{c}{\emph{Benign (Natural samples)}}\\
    \midrule
    Contrastive & {\textemdash} &            59.1 \pm 0.0 &                21.0 \pm 0.0 &  \highlight 74.0 \pm 0.0 &                20.9 \pm 0.0 &  \highlight 71.8 \pm 0.0 &      \highlight 44.7 \pm 0.0 &            78.2 \pm 0.0 &                73.6 \pm 0.0 \\
     Triplet & {\textemdash} &  \highlight 59.3 \pm 0.0 &      \highlight 21.7 \pm 0.0 &            74.0 \pm 0.0 &      \highlight 21.4 \pm 0.0 &            69.6 \pm 0.0 &                42.1 \pm 0.0 &  \highlight 81.9 \pm 0.0 &      \highlight 77.0 \pm 0.0 \\
    VisualPhishNet & {\textemdash} & \multicolumn{6}{p{\dimexpr26.2em+9\tabcolsep\relax}}{\Vhrulefill{}\text{N/A}\Vhrulefill{}} &            64.0 \pm 0.0 &                17.6 \pm 0.0 \\ \hlrow
      Contrastive ($\norm_{2}$) &  {\textemdash} &             55.6 \pm 0.0 &                 19.5 \pm 0.0 &             71.6 \pm 0.0 &                 18.8 \pm 0.0 &             66.3 \pm 0.0 &                 38.4 \pm 0.0 &             76.0 \pm 0.0 &                 70.9 \pm 0.0 \\ \hlrow
     Triplet ($\norm_{2}$) &  {\textemdash} &             55.9 \pm 0.0 &                 19.8 \pm 0.0 &             71.6 \pm 0.0 &                 18.8 \pm 0.0 &             62.2 \pm 0.0 &                 34.6 \pm 0.0 &             76.8 \pm 0.0 &                 73.8 \pm 0.0 \\ \hlrow
     Contrastive ($\norm_{\infty}$) &  {\textemdash} &             58.2 \pm 0.0 &                 20.2 \pm 0.0 &             72.1 \pm 0.0 &                 19.5 \pm 0.0 &             66.7 \pm 0.0 &                 39.0 \pm 0.0 &             76.8 \pm 0.0 &                 72.1 \pm 0.0 \\ \hlrow
      Triplet ($\norm_{\infty}$) &  {\textemdash} &             53.4 \pm 0.0 &                 17.9 \pm 0.0 &             71.9 \pm 0.0 &                 19.8 \pm 0.0 &             64.0 \pm 0.0 &                 36.4 \pm 0.0 &             79.1 \pm 0.0 &                 76.0 \pm 0.0 \\
    \midrule
    \multicolumn{10}{c}{$\norm_{2}(\epsilon = 4)$}\\
    \midrule
    Contrastive & PGD &             8.6 \pm 0.2 &                 4.8 \pm 0.1 &             3.7 \pm 0.2 &                 2.4 \pm 0.0 &             2.3 \pm 0.0 &                 2.3 \pm 0.0 &            45.5 \pm 0.2 &                43.1 \pm 0.1 \\
    Triplet & PGD &             9.6 \pm 0.3 &                 5.5 \pm 0.1 &             2.9 \pm 0.1 &                 2.2 \pm 0.0 &             1.2 \pm 0.0 &                 1.8 \pm 0.0 &            34.6 \pm 0.6 &                28.5 \pm 0.1 \\ \hlrow
    Contrastive ($\norm_{2}$) &   PGD &              26.4 \pm 0.3 &                  11.1 \pm 0.0 &     \highlight  37.2 \pm 0.2 &            \highlight   9.7 \pm 0.0 &   \highlight   51.6 \pm 0.0 &      \highlight    29.2 \pm 0.0 &      \highlight 56.6 \pm 0.4 &   \highlight 54.0 \pm 0.2 \\ \hlrow
    Triplet ($\norm_{2}$)  &   PGD &    \highlight  27.2 \pm 0.2 &   \highlight 11.3 \pm 0.1 &              37.0 \pm 0.4 &                   9.3 \pm 0.0 &              37.7 \pm 0.1 &                  20.3 \pm 0.0 &              55.1 \pm 0.1 &                  50.9 \pm 0.2 \\
    \midrule
    \multicolumn{10}{c}{$\norm_{\infty}(\epsilon = 0.01)$}\\
    \midrule
    Contrastive & PGD &             3.9 \pm 0.3 &                 3.4 \pm 0.1 &             1.4 \pm 0.2 &                 1.9 \pm 0.0 &             0.7 \pm 0.0 &                 1.5 \pm 0.0 &            35.7 \pm 0.2 &                34.3 \pm 0.2 \\
    Triplet & PGD &             4.0 \pm 0.3 &                 3.8 \pm 0.1 &             0.7 \pm 0.0 &                 1.7 \pm 0.0 &             0.2 \pm 0.0 &                 1.3 \pm 0.0 &            20.8 \pm 0.1 &                16.8 \pm 0.3 \\
    VisualPhishNet & PGD &  \multicolumn{6}{p{\dimexpr26.2em+9\tabcolsep\relax}}{\notapp} &            42.8 \pm 0.2 &                13.2 \pm 0.0 \\ \hlrow
     Contrastive ($\norm_{\infty}$) &   PGD &    \highlight 20.3 \pm 0.5 &        \highlight 8.8 \pm 0.0 &     35.7 \pm 0.2 &        \highlight 9.7 \pm 0.1 &  \highlight 53.6 \pm 0.0 &      \highlight 30.4 \pm 0.0 &    \highlight 56.7 \pm 0.2 &        \highlight 53.6 \pm 0.1 \\ \hlrow
  Triplet ($\norm_{\infty}$) &   PGD &              16.9 \pm 0.1 &                   7.4 \pm 0.1 &    \highlight 36.2 \pm 0.5 &                   9.6 \pm 0.1 &              39.3 \pm 0.1 &                  21.3 \pm 0.0 &              54.7 \pm 0.1 &                  49.1 \pm 0.1 \\
    \bottomrule
  \end{tabular}
  \vskip -0.1in
\end{table*}

Our experiments explore the following research questions:
\begin{questions}[itemsep=3pt]
\item How robust are \emph{naturally} trained \ac{DML} models towards established adversarial example attacks?\label{rq:attack}%
  \vspace{.5em}\\
  \textit{Among commonly used datasets for visual similarity, we find that \ac{DML} models, trained with state-of-the-art parameter choices, are vulnerable to adversarial examples, similar to non-DML models (Table~\ref{table:comparisons}). This forms our baseline for adversarial robustness.}

\item What is the accuracy of DML models when they are trained using our robust formulation?\label{rq:robust}%
  \vspace{.5em}\\
  \textit{We find that \ac{DML} models can be trained to become more robust across a variety of norms. For example, for a PGD attack with 5 iterations under $\norm_{\infty}(\epsilon = 0.01)$, we increase the adversarial accuracy to 53.6\% compared from the state-of-the-art natural baseline of 0.2\% for contrastive loss on the SOP dataset (Table~\ref{table:comparisons}).}

\item How does the robust training objective affect the learned embedding space?~\label{rq:effect}
  \vspace{.5em}\\
  \textit{Using a synthetic dataset, we demonstrate that the proposed adversarial training reduces the amount of shifting that adversarial perturbations can cause in the embedding space (Figure~\ref{fig:gaussian}).}

\end{questions}

We run all experiments on Nvidia Tesla V100 GPUs (32 GB) RAM. Our code is available at \urlrepository{}.

\subsection{Experimental Setup}


\paragraph{Datasets.}
We use the following four real-world image datasets for our experiments:
\begin{itemize}[itemsep=3pt]
\item \textbf{CUB200-2011}~\citep{welinder10}: Images of birds across \num{200} species and have a total of \num{11788} images.
\item \textbf{CARS196}~\citep{krause13}: Dataset with images of cars spanning across $196$ models, with a total of \num{16185} images.
\item \textbf{SOP}~\citep{song16}: Product images from eBay listings \num{120053} images of \num{22634} different online products.
\item \textbf{VisualPhish}~\citep{abdelnabi20}: Screenshots of benign websites, from a set of established brands, and phishing websites that attempt to replicate the visual appearance of their benign counterpart.
  It covers \num{146} brands across a total of \num{10558} screenshots.
\end{itemize}

CUB200-2011, CARS196, and SOP are commonly used within the \ac{DML} literature~\cite{musgrave20}.
These three datasets are divided into a training and testing set of approximately the same size by selecting the first half of classes for the training set, while having the remaining classes be in the testing set~\cite{roth20}. This setup reflects an out-of-distribution scenario --- a common application of \ac{DML}. VisualPhish is a newer dataset that underlies the robust phishing detection model, VisualPhishNet~\cite{abdelnabi20}. For a fair comparison, we adopt the train-test split from the VisualPhish implementation. This yields a test set of \num{717} website screenshots. In addition to these real-world datasets, we also include the following synthetic dataset:

\noindent\textbf{Synth Dataset:} A dataset with two classes \emph{a} and \emph{b} where data points $\datapoint \in [0,1]^{k}$ and $k = 224 \times 224 \times 3$ to maintain identical dimensionality of the real-world datasets.
  Data points from each class are drawn from $\mathcal{N}(\mu, \sigma^{2}I)$ st. ${\sigma = 0.075}$ while ${\mu = 0.25}$ for class \emph{a} and ${\mu = 0.75}$ for class \emph{b}.

\myparagraph{Models \& Training Parameters.}
We use default parameter choices from prior work that yield state-of-the-art performance on natural samples~\cite{roth20}.
Main parameters are summarized in this section and provide a complete listing in Appendix~\ref{app:params}.
Deviations from the default parameter choices are discussed and emphasized.

All models are ResNet50~\cite{he16} initialized with pre-trained weights from an ImageNet classifier.
We replace the last fully connected layer with another that matches the embedding space dimensionality.
Embeddings are normalized to be on the $n$-dimensional unit sphere, where $n = \num{128}$ throughout our experiments.
We use ADAM~\cite{kingma15} with learning rate\footnote{This learning rate differs from the one stated by \citet{roth20} in their publication, $10^{-5}$, but reflects the actual learning rate used throughout their experiments. See the field ``lr'' within experiment configuration: https://bit.ly/3a4FyHP.} of $10^{-6}$, weight decay of $4 \cdot 10^{-4}$.
We use contrastive and triplet losses during training, setting $\alpha = 1.0$ and $\alpha = 0.2$, respectively.




To the best of our knowledge, VisualPhishNet is the only previous attempt at creating an adversarially robust \ac{DML} model trained using metric losses.
At the core, the model is a variant of the VGG16~\cite{simonyan15} architecture with an unnormalized embedding layer of size 512.
It was trained using the VisualPhish dataset and is expected to learn a visual similarity metric between web sites of various origins.

Training on the real-world datasets is performed over 150 epochs, with the exception of SOP that is trained for 100 epochs due to its volume~\cite{roth20}.
Mini-batches are of size 112 and sampled using the sampling technique SPC-2, which ensures that each batch contains exactly two samples per class for the selected classes in the batch.

\myparagraph{Adversarial Robustness.}
To establish a benchmark for adversarial robustness, we employ the attack algorithm covered in Section~\ref{sec:attackalg}.
For each data point being perturbed, we sample the nearest positive neighbor to reflect the ideal attack setting for an adversary.
The formulation uses \ac{PGD}~\cite{madry18} because it is considered one of the strongest white-box attacks available~\cite{wong20}. Each attack is run for five iterations ($i = 5$) and has a step size given by $2\epsilon\frac{1}{i}$, such that the step size remains small while not hindering the optimization from reaching any point within the $\epsilon$-ball despite random initialization.
Throughout the experiments we use the notation of $\norm_{p}(\epsilon = 0.01)$ to indicate that for any data point $\datapoint$, its valid perturbations are contained in  $B_{p}(\datapoint, 0.01)$.
We compute the adversarial robustness for $\norm_{\infty}(\epsilon = 0.01)$ to accommodate VisualPhishNet~\cite{abdelnabi20}, and  $\norm_{2}(\epsilon = 4)$ to provide comparisons for an alternative norm.
In addition to \ac{PGD}, we also investigate adversarial robustness towards the \ac{CW} attack algorithm~\cite{carlini17}, which can be found in Appendix~\ref{app:cw}.

\myparagraph{Adversarial Training.}
Given that natural training of \ac{DML} models is already considered an expensive procedure~\citep{roth20}, solving the inner-maximization of the proposed robust formulations in Section~\ref{sec:robustdml} can make the procedure even more expensive and potentially infeasible for practical applications.
As previously discussed, the inner-maximization is solvable using traditional first-order attack methods, e.g. FGSM, \ac{PGD}, and \ac{CW}.
This fact enables us to apply a training technique by \citet{wong20}, that involves adversarial training using the cheaper R+FGSM~\cite{tramer18} attack, in conjunction with early-stopping.
This yields similar increases in robustness towards stronger and more expensive attacks, such as \ac{PGD}, despite not being directed trained on these attacks.
For this attack, we define $\alpha = \epsilon \cdot 0.25$ as we have empirically determined that it is effective and training \ac{DML} models. Using the proposed attack algorithm for adversarial training (Section~\ref{sec:advtrain}), we perturb the positive data points.
This choice was to avoid affecting the relative distances to negative data points, which can induce instabilities during the training of \ac{DML} models if they become too small~\cite{wu17}.

\myparagraph{Evaluation Metrics.}
To evaluate the performance of the trained models, we employ the following \ac{DML}-specific evaluation metrics: Recall at One (R@1) and Mean Average Precision at R (mAP@R)~\cite{musgrave20}.
R@1 is effectively the accuracy of class inference using the class of the nearest neighboring anchor within the embedding space produced by the model.
Given the test set ${\dataset = \{(\mathbf{x}_{1},y_{1}),\cdots, (\mathbf{x}_{n}, y_{n}) \}}$, and the function $I_{k}(i)$ that outputs the indices of the $k$-nearest neighbors for a data point $\datapoint_{i}$, such that
\begin{gather}
  I_{k}(i) = \argmin_{\substack{\vert K \vert = k \\ i \notin K}}\sum_{j \in K} \distfunc(\datapoint_{i}, \datapoint_{j}) \; ,
\end{gather}
then R@1 given by:
\begin{gather}
  \text{R@}1 = \frac{1}{n} \sum_{\substack{i \in \{1,\cdots,n\} \\ j \in I_{1}(i)}} 1_{y_{i}=y_{j}} \; .
\end{gather}

mAP@R is metric for measuring a model's ability to rank classes in the embedding space; we adopted this metric for the reasons covered by \citet{musgrave20}.
It is defined as

\begin{gather}
    \text{mAP@R} = \frac{1}{n}\sum^{n}_{i=1} \frac{1}{R_{i}}\sum^{R_{i}}_{k=1} \frac{1}{k} \sum_{j \in I_{k}(i)} 1_{y_{i}=y_{j}} \; ,
  \end{gather}

where $R_{i} = \sum_{j \in \{1,\cdots,n\} \setminus \{i\}} 1_{y_{i}=y_{j}}$.

\subsection{Experimental Results}\label{sec:expres}

\begin{figure}[h!]
  \centering
  \includegraphics[width=\linewidth]{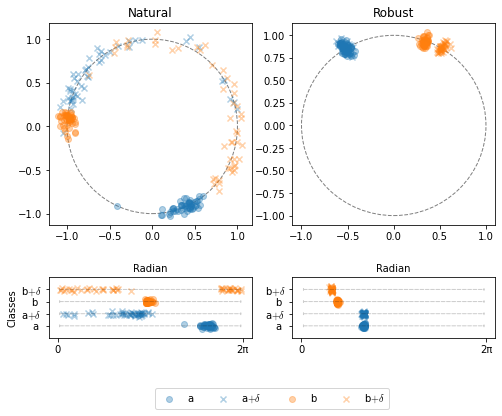}
    \caption{
      Embedding spaces of two trained models.
      (first row) Visualization of the embedding spaces of the two models (natural and robust).
      Effectively the embeddings are normalized to the unit circle, however, small distortions are added to improve visual clarity.
      Circles mark embeddings of benign data points, while crosses are data points with an adversarial perturbation ($\delta$).
      (second row) Alternative visualization of the embedding space, showing the point relative to their radian on isolated spheres.
      It can be seen that embeddings of the robustly-trained model shifts much less, when faced with adversarial perturbed input, and are thus more robust.~\label{fig:synthetic}
  }
  \label{fig:gaussian}
\end{figure}

\myparagraph{Robustness of Natural Training~(\ref{rq:attack})}
We establish a baseline of robustness against adversarial perturbations for naturally-trained \ac{DML} models across the covered metric losses, $\norm_{p}$-norms, and datasets.
Results can be seen in Table~\ref{table:comparisons}.
Across any of the common real-world datasets (CUB200-2011, CARS196, SOP) it can be seen that both the model's ability to infer the correct class from its nearest neighbor (R@1) and its ability to rank classes (mAP@R) drops by several orders of magnitude.
Exemplifying this, the naturally-trained model using triplet loss on CUB200-2011 drops from $59.3\%$ accuracy (on benign data) down to $4.0\%$ (on adversarially-perturbed data) for $\norm_{\infty}(\epsilon = 0.01)$.
Naturally-trained models on CARS196 and SOP yield comparable or worse adversarial robustness.
Notably, naturally-trained models on the VisualPhish dataset achieves a higher baseline for adversarial robustness.
We suspect this deviation, from the other common real-world datasets, is related to the underlying data distribution of the dataset.

We conclude that naturally-trained \ac{DML} models are not inherently robust, contrary to what results of prior work might indicate~\citep{abdelnabi20}.
We suspect this difference might stem from the method of attack or the fact we use a stronger first-order attack.

\myparagraph{Adversarial Training for Robustness~(\ref{rq:robust})}
From Table~\ref{table:comparisons}, it can be seen that the proposed method for adversarial training increases adversarial robustness (accuracy and ability to rank) across the chosen metric losses, norms, attacks and datasets.
As an example, the robust R@1 on SOP increases from $0.7\%$ (naturally-trained) to $56.3\%$ for $\norm(\epsilon=0.01)$.
We also observe that the proposed method increases the adversarial robustness (in terms of R@1) on the VisualPhish dataset to $56.7\%$, and thus outperforms the prior work of VisualPhishNet~\cite{abdelnabi20}, which achieves $42.8\%$.
As shown in Appendix~\ref{app:cw}, it can be seen that the gained robustness also applies to alternative attacks (\ac{CW}) for $\norm_{2}(\epsilon = 4)$.
Performance of the trained robust models on benign input remains largely unaffected.
Figure~\ref{fig:exampleinfer} shows an example of inference under different training objectives, and we provide a publicly available gallery of other examples\footnote{\urlgallery{}}.


\myparagraph{Effects on Embedding Space~(\ref{rq:effect})}
Using the described synthetic dataset, we investigate the effect of adversarial training on the learned embedding space.
The experiment involves training a \ac{DML} model to map data points of the high-dimensional synthetic dataset, with the classes \emph{a} and \emph{b}, onto to a two-dimensional embedding space.
We choose to have the embedding space be two-dimensional to allow visualizations of the learned embedding space.
Each of the models, naturally-trained and robust, uses contrastive loss and is trained on approximately \num{15}K data points.
Adversarial perturbations are derived using the proposed attack formulation with \ac{PGD} under $\norm_{\infty}(\epsilon = 0.01)$.
Differences of the learned embedding spaces, and the influence of the adversarial perturbations, is shown in Figure~\ref{fig:synthetic}.
We observe that the robust model is capable of maintaining smaller inter-class distances between adversarially perturbed data points and benign data points.

\section{Conclusion}
Deep Metric Learning (DML) creates feature embedding spaces where similar input points are geometrically close to each other, while dissimilar points are far apart.
However, the underlying DNNs are vulnerable to adversarial inputs, thus making the DML models themselves vulnerable.
We demonstrate that naturally-trained DML models are vulnerable to strong attackers, similar to other types of deep learning models.
To create robust DML models, we contribute a robust training objective that can account for the \emph{dependence} of metric losses --- the phenomenon that the loss at any point depends on the other items in the mini-batch and the sampling process that was used to derive the mini-batch. Our robust training formulation yields robust DML models that can withstand PGD attacks without severely degrading their performance on benign inputs.

\bibliography{bibliography}
\bibliographystyle{icml2021}

\appendix
\clearpage
\section{Training Parameters (Expanded)}\label{app:params}
This section expands upon details and hyper-parameters used throughout the training of the respective \ac{DML} models.

\paragraph{Batches \& Sampling.}
Recall, that the training process uses a mini-batch size of 112 data points.
Each mini-batch is sampled such that it contains exactly two samples per class~\citep{roth20}.
Following this, sets of tuples or triplets are derived (depended on loss) from the mini-batch using distance weighted sampling~\cite{wu17} for negatives, while positives are given by pair-based sampling.
Distance weighted sampling enhances the stability of training using metric losses, that can suffer from becoming stuck at a local minima early on in the training procedure~\cite{wu17}.
The cardinality of the triplet-set is identical to the mini-batch size.
The size of the tuple-set is double the size of the mini-batch, thus balancing out the number of data points being compared relative to the triplet-set.
Furthermore, each data point within the tuple-set is used in a positive and negative pair.

\paragraph{Data Augmentation.} We augment the dataset using the following operations for each input image: (1) random cropping to an image patch of size 60-100\% of the original image area; (2) scaling; (3) normalization of pixel intensities. One difference is that our patch sizes differ from \citet{roth20} that employs patches of size 8-100\% of original area. We change this parameter because recent work suggests that computer vision models can be biased by backgrounds and textures during during~\citep{xiao20}. To combat this, we use cropping and scaling values based on \citet{szegedy15}.

\section{Alternative Attack (Carlini-Wagner)}~\label{app:cw}
The Carlini-Wagner (\ac{CW}) attack is an unbounded attack, and thus constrains perturbations to lie within the given $\norm_{p}$~\cite{carlini17}.
We employ a clipping technique similar to \citet{tramer19}, which projects the perturbation to the $\norm_{p}$-ball at every step.
Additionally, as inference is costly for \ac{DML} models (nearest neighbor search across embedding space), the ability of providing early stopping mechanism has been disabled.
Results are presented in Table~\ref{table:cw}. This is our best effort on providing strong hyper-parameters for the attack.
It can be seen that the robustly trained model manages to remain higher robustness towards \ac{CW} attacks than the stronger \ac{PGD} attack.
The impact of the mentioned alterations, and the used hyper-parameters could yield the \ac{CW} attack to be non-optimal.
Thereby, these results should be seen as an lower-bound representation of robustness towards the \ac{CW}, despite \ac{PGD} generally being consider the state-of-the-art~\cite{wong20}.

\begin{table*}[h!]
  \caption{
    Performance of robust \ac{DML} models over four datasets for $\norm_{2}(\epsilon = 4)$ against the \ac{CW} attack algorithms.
    Results are an average of five random seeds.
    Recall that, R@1 reflects a model's inference accuracy, while mAP@R reflects its ability to rank similar entities.
    Our robust training objective also provides robustness towards \ac{CW} attacks.
  }
  \label{table:cw}
  \vskip 0.15in
  \setlength{\tabcolsep}{4pt}
  \sisetup{detect-all = true}
  \small
  \centering
  \begin{tabular}{
    l
    l
    *{4}{
    @{\hskip 2\tabcolsep}
    S[table-format=2.2(2),table-column-width=4.4em,table-align-uncertainty = false]
    S[table-format=2.2(2),table-column-width=4.4em,table-align-uncertainty = false]
    }}
    \toprule
    & & \multicolumn{2}{c}{\textbf{CUB200-2011}} & \multicolumn{2}{@{}c@{\hskip 4\tabcolsep}}{\textbf{CARS196}} & \multicolumn{2}{@{}c@{}}{\textbf{SOP}} & \multicolumn{2}{@{}c@{}}{\textbf{VisualPhish}} \\
    \cmidrule(lr{\dimexpr 4\tabcolsep-0.5em}){3-4} \cmidrule(l{-0.5em}r{\dimexpr 4\tabcolsep-0.5em}){5-6} \cmidrule(l{-0.5em}r{\dimexpr 4\tabcolsep-0.5em}){7-8} \cmidrule(l{-0.5em}r{0.5em}){9-10}
    Model & Attack & {R@1} & {mAP@R} & {R@1} & {mAP@R} & {R@1} & {mAP@R} & {R@1} & {mAP@R}\\
    \midrule
    \multicolumn{10}{c}{$\norm_{2}(\epsilon = 4)$}\\
    \midrule
    Contrastive ($\norm_{2}$) &   PGD &              26.4 \pm 0.3 &                  11.1 \pm 0.0 &       37.2 \pm 0.2 &               9.7 \pm 0.0 &      51.6 \pm 0.0 &          29.2 \pm 0.0 &       56.6 \pm 0.4 &    54.0 \pm 0.2 \\
    \dittoarrow  &    \bfseries \ac{CW} &    \bfseries           40.6 \pm 0.4 &    \bfseries               17.2 \pm 0.1 &    \bfseries           58.7 \pm 0.2 &    \bfseries      16.4 \pm 0.0 &    \bfseries  63.2 \pm 0.0 &    \bfseries      37.2 \pm 0.0 &    \bfseries           67.8 \pm 0.2 &    \bfseries  64.6 \pm 0.1 \\
    Triplet ($\norm_{2}$)  &   PGD &      27.2 \pm 0.2 &    11.3 \pm 0.1 &              37.0 \pm 0.4 &                   9.3 \pm 0.0 &              37.7 \pm 0.1 &                  20.3 \pm 0.0 &              55.1 \pm 0.1 &                  50.9 \pm 0.2 \\
    \dittoarrow &    \bfseries \ac{CW} &    \bfseries 42.0 \pm 0.1 &    \bfseries 17.7 \pm 0.1 &    \bfseries 59.6 \pm 0.1 &    \bfseries 16.4 \pm 0.0 &    \bfseries  54.5 \pm 0.0 &    \bfseries 30.5 \pm 0.0 &    \bfseries  68.7 \pm 0.2 &    \bfseries  65.8 \pm 0.2 \\
    \bottomrule
  \end{tabular}
  \vskip -0.1in
\end{table*}
\section{Theoretical Analysis}

\subsection{Robustness and Lipschitzness of DML}
In Section~\ref{sec:intro}, we pointed out that the Lipschitzness of the DML model also plays
an important role as in the traditional classifier situation. Here we have a formal analysis.

Let the sample space $\calX$ be $\bigcup_{y \in \calY} \calX_y$, where
$\calY = \{ y_1,\cdots,y_L \}$ is the space of labels and $\calX_y
\subseteq \calX$ is the set of samples with label $y$. Suppose we have
a deep embedding model $f_\theta$ with parameter $\theta$ trained using
one of the loss functions described earlier.  Let $A = \{
(\bfa_1,c_1),\cdots, (\bfa_m,c_m)\}$ be a reference dataset (e.g. a set
of faces along with their label), which we call {\it anchors}.
Suppose we have a sample $\bfz$ and let $k (A,\bfz)$ be the index that
corresponds to $\argmin_{j \in \{1,\cdots,m \} } d_\theta
(\bfa_j,\bfz)$. We predict the label of $\bfz$ as $lb(A,\bfz) \; = \;
c_{k(A,\bfz)}$.  Recall that $d_\theta(\bfx,\bfz)$ is the
distance metric in the embedding space $\parallel f_\theta(\bfx) -
f_\theta (\bfz) \parallel_2$.

We will assume that our sample space $\calX$ is a metric space with metric $\mu$.
A  point $\bfz \in \calX$ is {\it $\epsilon$-robust} w.r.t. $A$, $\mu$ and $d_\theta$ iff
for $k (A,\bfz) = \argmin_{1 \leq i \leq m} d_\theta (a_i,\bfz)$, we have that
for all $j \not= k (A,\bfz)$ and $\mu(\bfx,\bfz) \leq \epsilon$, $d_\theta(a_j,\bfx) > d_\theta (a_{k_\mu (A,\bfz)},\bfx)$.
In other words, perturbing $\bfz$ by $\epsilon$ in the sample space does not change the anchor
it is close to in the embedding space.

A  point $\bfz \in \calX$ is {\it $\delta$-separated} w.r.t. $A$ and $d_\theta$ iff
for $k (A,\bfz) = \argmin_{1 \leq i \leq m} d_\theta (a_i,\bfz)$ we have that
for all $j \not= k (A,\bfz)$, $d_\theta (a_j,\bfz) > d_\theta (a_{k_\mu (A,\bfz)},\bfz) + \delta$.
In other words, $f_\theta(\bfz)$ is at least $\delta$ closer to its anchor than 
other anchors in the embedding space. As a result, $f_\theta$ correctly classifies $\bfz$.


We assume that $d_\theta$ is $L$-Lipschitz, i.e., for all $\bfx$ and $\bfz$ in $\calX$:
\begin{eqnarray*}
  d_\theta(\bfx,\bfz) \leq & L \mu (\bfx,\bfz)
\end{eqnarray*}

\begin{lemma}
  \rm
  If $L\leq \delta/(2\epsilon)$, and $\bfz$ is $\delta$-separated w.r.t. $A$ and $d_\theta$, then 
  $\bfz$ is $\epsilon$-robust.
\end{lemma}
\newcommand{\dist}{d_\theta}

\begin{proof}
Let $i = k (A,\bfx)$, $j\neq i$, and $\mu(\bfx, \bfz)\leq \epsilon$.
\begin{eqnarray*}
  \dist(a_i, \bfx) &\leq &\dist(\bfx, \bfz) + \dist(\bfz, a_i)\\
                    &\leq& L\mu(\bfx, \bfz) + \dist(\bfz, a_i)\\
                    & < &L\mu(\bfx, \bfz) + \dist(\bfz, a_j)-\delta \\
                    &\leq& L\mu(\bfx, \bfz) + [\dist(\bfz, \bfx)+ \dist(\bfx, a_j)]-\delta\\
                    &\leq& L\mu(\bfx, \bfz) + L\mu(\bfz, \bfx)+ \dist(\bfx, a_j)-\delta\\
\end{eqnarray*}
Because $L\leq \delta/(2\epsilon)$, $\mu(\bfz, \bfx)\leq \epsilon$, we have
\[
    \dist(a_j,\bfx) > \dist(a_i, \bfx) + (\delta-2L\epsilon) \geq \dist(a_i, \bfx),
\]
so
\[
    \dist(a_i, \bfx) < \dist(a_j, \bfx).
\]
\end{proof}

\newcommand{\RR}{\mathbb{R}}

\subsection{DML with Gaussian Mixture Model}
 To further motivate the connection
 between robustness of an embedding and its Lipschitz constant, we
 consider a Gaussian mixture model. These models have been considered
 in the theoretical analysis of robustness in the classification
 setting~\cite{dan:2020,tsipras:2019}. Our synthetic dataset experiment (Figure~\ref{fig:gaussian})
 illustrates this Gaussian mixture model setting. Let $\mathcal{N}(\mu,\Sigma)$
 be the Gaussian distribution in $\RR^n$ with mean $\mu \in \RR^n$ and
 $\Sigma$ a $n \times n$ positive-definite matrix. We will consider
 Gaussian distributions of the form $\mathcal{N}(\mu,I_n)$ where $I_n$
 is the $n \times n$ identity matrix.

 Let $\calX \times \calY$ (where $\calY = \{ -1 , 1 \}$) be generated
 from a distribution $\calD$ as follows: $y \in \calY$ is equally probable
 with probability $\frac{1}{2}$ and given $y$, generate $\bfx$ according to
 $\mathcal{N}(y\mu,I_n)$.

 We have the following concentration of measure result from Theorem 5.2.2~\cite{vershynin}.
 \begin{theorem}
 \label{thm:concentration}
   \rm (Gaussian concentration) Consider a random vector $\bfx \sim \mathcal{N}(0,I_n)$ and
   a Lipschitz function $f: \RR^n \rightarrow \RR$. Then
   \begin{eqnarray*}
     \parallel f(\bfz) - E f(\bfx) \parallel_{\psi_2} & \leq & C \parallel f \parallel_{Lip},
   \end{eqnarray*}
   where $\parallel f \parallel_{Lip}$ is the Lipschitz constant of $f$, and $\parallel \cdot \parallel_{\psi_2}$ is
   the sub-Gaussian metric.
 \end{theorem}

Consider a DML model $f_\theta: \RR^n \rightarrow S^d$, and let
$d_\theta$ be the associated distance metric. Let $a_1$ and $a_{-1}$
be the anchors for labels $1$ and $-1$ respectively. Consider the two
functions defined as follows: $f_1 (\bfx) = d_\theta (a_1,\bfx)$ and
$f_{-1} (\bfx) = d_\theta (a_{-1},\bfx)$ (the functions correspond
to the distances from the two anchors). 
\begin{eqnarray*}
  \beta_1 & = & E_{\bfx \sim \mathcal{N}(\mu,I_n)} f_1 (\bfx) \\
  \beta_{-1} & = & E_{\bfx \sim \mathcal{N}(-\mu,I_n)} f_2 (\bfx)
\end{eqnarray*}

We first show that $f_1$ is $L$-Lipschitz if $f_\theta$ is $L$-Lipschitz.
Take $\bfx, \bfz \in \calX$, 
\begin{eqnarray*}
  |f_1(\bfx) - f_1(\bfz)| & = & |\dist(a_1,\bfx) - \dist(a_1, \bfz)|\\
  &\leq& \dist(\bfx, \bfz)\\
  &\leq& L\mu(\bfx, \bfz)\\
\end{eqnarray*}

As a result, $\parallel f_1 \parallel_{Lip} = \parallel f_\theta \parallel_{Lip}$.
Intuitively, if the Lipschitz constant of $f_1$ is lower, the points drawn
from $\mathcal{N}(\mu,I_n)$ get closer to $\beta_1$. In other words, as
the Lipschitz constant of embedding gets smaller, the ``point clouds''
corresponding to the two Gaussian distributions in the mixture get
farther apart, because they are concentrated more around their means. 

Next we formalize this intuition. Let $E(\bfx,a_1,a_{-1})$ represent
the event that $\bfx$ is closer to $a_{-1}$ than $a_1$. We prove the
following:
\begin{equation}\label{eq:ineq}
 P_{\bfx \sim \mathcal{N}(\mu,I_n)} (1_{E(\bfx,a_1,a_{-1})})  \leq  2 \exp \left( - \frac{C' z^2}{\parallel f_1 \parallel_{Lip} } \right)
\end{equation}
In the equation given above, $C' >0$ is a positive constant, 
and $z$ is given by the following expression:
\[
\frac{d_\theta(a_{-1},a_{1})}{2} - \beta_1
\]
Notice that $P_{\bfx \sim \mathcal{N}(\mu,I_n)} (1_{E(\bfx,a_1,a_{-1})})$ represents the probability that a point
drawn from $\mathcal{N}(\mu,I_n)$ is closer to $a_{-1}$ than $a_{1}$, and hence represents an ``undesirable event''.
Also note that the upper bound goes down as the Lipschitz constant $\parallel f_1 \parallel_{Lip}$ goes down, and
thus confirming our intuition. Next we prove Equation~\ref{eq:ineq}.

Let $X$ be a sub-Gaussian random variable, then the following equation is well-known:
\begin{eqnarray}
\label{eqn:sub-guassian}
P(\mid X \mid \geq t) & \leq & 2 \exp \left( \frac{-c t^2}{\parallel X \parallel^2_{\psi_2}} \right)
\end{eqnarray}

To prove the Equation~\ref{eq:ineq}, we use the following sequence of inequalities (let
$q = P_{\bfx \sim \mathcal{N}(\mu,I_n)} (1_{E(\bfx,a_1,a_{-1})})$)
\begin{eqnarray*}
  q & \leq &
  P_{\bfx \sim \mathcal{N}(\mu,I_n)} \left( f_1(\bfx) \geq \frac{d_\theta(a_{-1},a_{1})}{2} \right) \\
  & \leq & P_{\bfx \sim \mathcal{N}(\mu,I_n)} \left(\mid f_1(\bfx) - \beta_1 \mid \geq  \frac{d_\theta(a_{-1},a_{1})}{2}-\beta_1 \right)\\
  & \leq & 2 \exp \left( - \frac{C' z^2}{\parallel f_1 \parallel_{Lip}} \right)
\end{eqnarray*}
The first step follows from the following observation: if $f_1(\bfx)$ is less than $\frac{d_\theta(a_{-1},a_{1})}{2}$
then $\bfx$ is closer to $a_1$ than $a_{-1}$. The next two steps use Theorem~\ref{thm:concentration} and
Equation~\ref{eqn:sub-guassian}.

\end{document}